\documentclass[10pt,twocolumn,letterpaper]{article}

\usepackage{cvpr}              

\usepackage[dvipsnames]{xcolor}

\definecolor{cvprblue}{rgb}{0.21,0.49,0.74}
\usepackage[pagebackref,breaklinks,colorlinks,citecolor=cvprblue]{hyperref}
\usepackage[accsupp]{axessibility} 

\definecolor{palered}{rgb}{0.914, 0.598, 0.598}
\definecolor{palepurple}{rgb}{0.555, 0.484, 0.762}

\def\paperID{1114} 
\def\confName{CVPR}
\def\confYear{2024}

\title{HyperDreamBooth: HyperNetworks for \\ Fast Personalization of Text-to-Image Models}

\author{
  Nataniel Ruiz ~~~~~~~ Yuanzhen Li ~~~~~~~ Varun Jampani ~~~~~~~ Wei Wei ~~~~~~~ Tingbo Hou \\ ~~ Yael Pritch ~~~~~~~ Neal Wadhwa ~~~~~~~ Michael Rubinstein ~~~~~~~ Kfir Aberman\\
  Google Research\\
}

\begin{document}

\twocolumn[{
\renewcommand\twocolumn[1][]{#1}
\maketitle
\begin{center}
    \centering
    \vspace*{-.5cm}
    \includegraphics[width=0.82\textwidth]{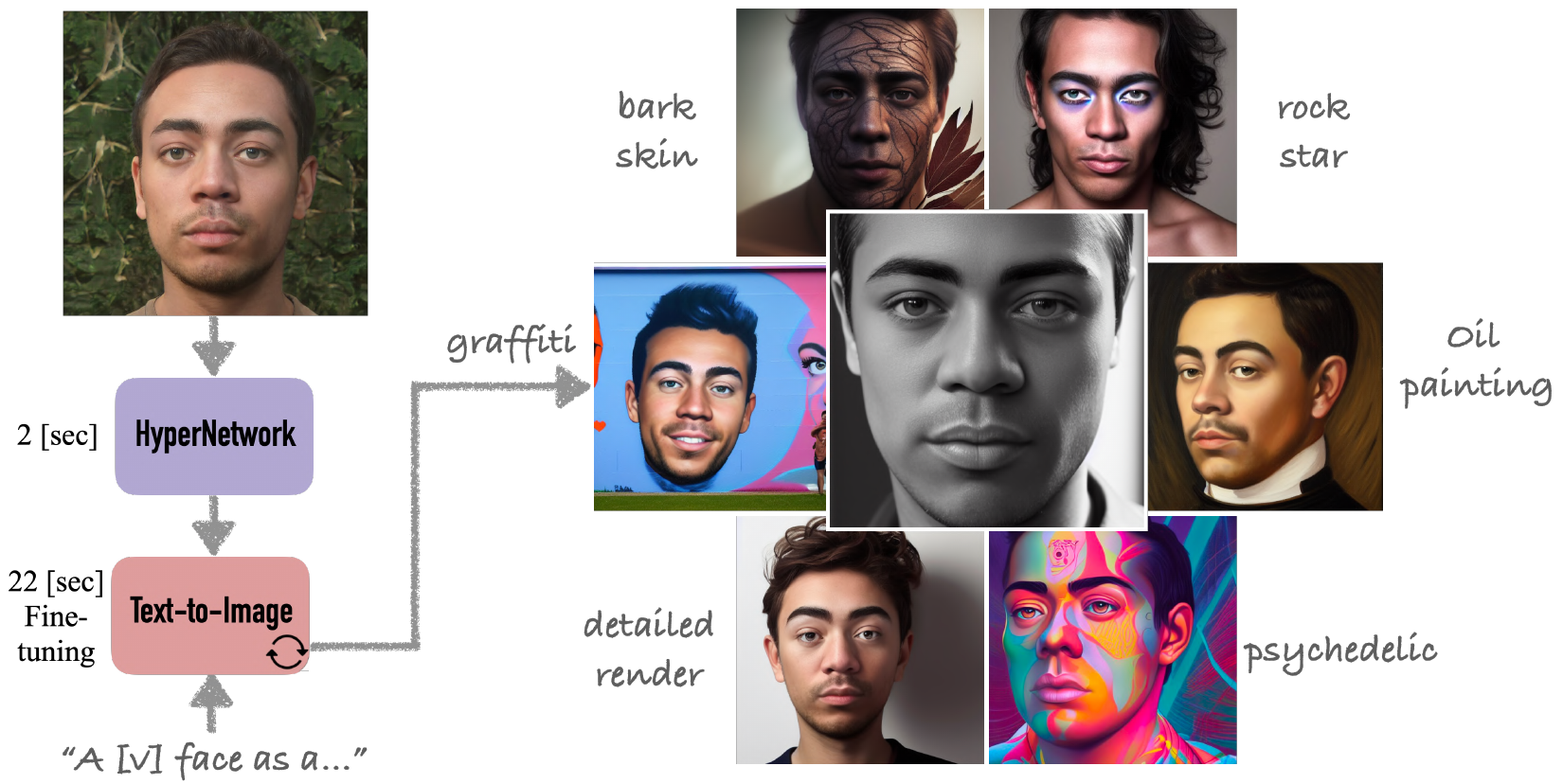}
    \vspace*{-.2cm}
    \captionof{figure}{Using only a \textit{single} input image, \textit{HyperDreamBooth} is able to personalize a text-to-image diffusion model \textbf{25x} faster than DreamBooth~\cite{ruiz2022dreambooth}, by using \textbf{\textcolor{palepurple}{(1)}} a HyperNetwork to generate an initial prediction of a subset of network weights that are then \textbf{\textcolor{palered}{(2)}} refined using fast finetuning for high fidelity to subject detail. Our method both \textit{conserves model integrity and style diversity} while closely approximating the subject's essence and details.}
\label{fig:teaser}
\end{center}
}]

\begin{abstract}
\vspace{-0.1in}
Personalization has emerged as a prominent aspect within the field of generative AI, enabling the synthesis of individuals in diverse contexts and styles, while retaining high-fidelity to their identities. However, the process of personalization presents inherent challenges in terms of time and memory requirements. Fine-tuning each personalized model needs considerable GPU time investment, and storing a personalized model per subject can be demanding in terms of storage capacity.
To overcome these challenges, we propose HyperDreamBooth—a hypernetwork capable of efficiently generating a small set of personalized weights from a single image of a person. By composing these weights into the diffusion model, coupled with fast finetuning, HyperDreamBooth can generate a person's face in various contexts and styles, with high subject details while also preserving the model's crucial knowledge of diverse styles and semantic modifications.
Our method achieves personalization on faces in roughly 20 seconds, \textbf{25x} faster than DreamBooth and \textbf{125x} faster than Textual Inversion, using as few as \textit{one} reference image, with the same quality and style diversity as DreamBooth. Also our method yields a model that is \textbf{10,000x} smaller than a normal DreamBooth model.
\vspace{-0.2in}
\end{abstract}
\vspace{-0.2in}
\section{Introduction}
\label{sec:intro}

Recent work on text-to-image (T2I) personalization~\cite{ruiz2022dreambooth} has opened the door for a new class of creative applications. Specifically, for face personalization, it allows generation of new images of a specific face or person in different styles. The impressive diversity of styles is owed to the strong prior of pre-trained diffusion model, and one of the key properties of works such as DreamBooth~\cite{ruiz2022dreambooth}, is the ability to implant a new subject into the model without damaging the model's prior. Another key feature of this type of method is that subject's essence and details are conserved even when applying vastly different styles. For example, when training on photographs of a person's face, one is able to generate new images of that person in animated cartoon styles, where a part of that person's essence is preserved and represented in the animated cartoon figure - suggesting some amount of visual semantic understanding in the diffusion model. These are two core characteristics of DreamBooth and related methods, that we would like to leave untouched. Nevertheless, DreamBooth has some shortcomings: size and speed. For size, the original DreamBooth paper finetunes all of the weights of the UNet and Text Encoder of the diffusion model, which amount to more than 1GB for Stable Diffusion. In terms of speed, notwithstanding inference speed issues of diffusion models, training a DreamBooth model takes about 5 minutes for Stable Diffusion (1,000 iterations of training). This limits the potential impact of the work. In this work, we want to address these shortcomings, without altering the impressive key properties of DreamBooth, namely \textit{style diversity} and \textit{subject fidelity}, as depctied in Figure~\ref{fig:teaser}. Specifically, we want to \textit{conserve model integrity} and \textit{closely approximate subject essence} in a fast manner with a small model.

Our work proposes to tackle the problems of \textbf{size} and \textbf{speed} of DreamBooth, while preserving \textbf{model integrity}, \textbf{editability} and \textbf{subject fidelity}. We propose the following contributions:
\begin{itemize}
    \item \textit{Lighweight DreamBooth (LiDB)} - a personalized text-to-image model, where the customized part is roughly 100KB of size. This is achieved by training a DreamBooth model in a low-dimensional weight-space generated by a random orthogonal incomplete basis inside of a low-rank adaptation~\cite{hu2021lora} weight space.
    \item New \textit{HyperNetwork} architecture that leverages the Lightweight DreamBooth configuration and generates the customized part of the weights for a given subject in a text-to-image diffusion model. These provide a strong directional initialization that allows us to further finetune the model in order to achieve strong subject fidelity within a few iteration. Our method is \textbf{25x} faster than DreamBooth while achieving similar performances.
    \item We propose the technique of \textit{rank-relaxed finetuning}, where the rank of a LoRA DreamBooth model is relaxed during optimization in order to achieve higher subject fidelity, allowing us to initialize the personalized model with an initial approximation using our HyperNetwork, and then approximate the high-level subject details using rank-relaxed finetuning.
\end{itemize}

One key aspect that leads us to investigate a HyperNetwork approach is the realization that in order to be able to synthesize specific subjects with high fidelity, using a given generative model, we have to ``modify" its output domain, and insert knowledge about the subject into the model, namely by modifying the network weights.

\begin{figure}[t]
    \hspace*{-0.05\columnwidth}
    \centering
    \includegraphics[width=1.1\columnwidth]{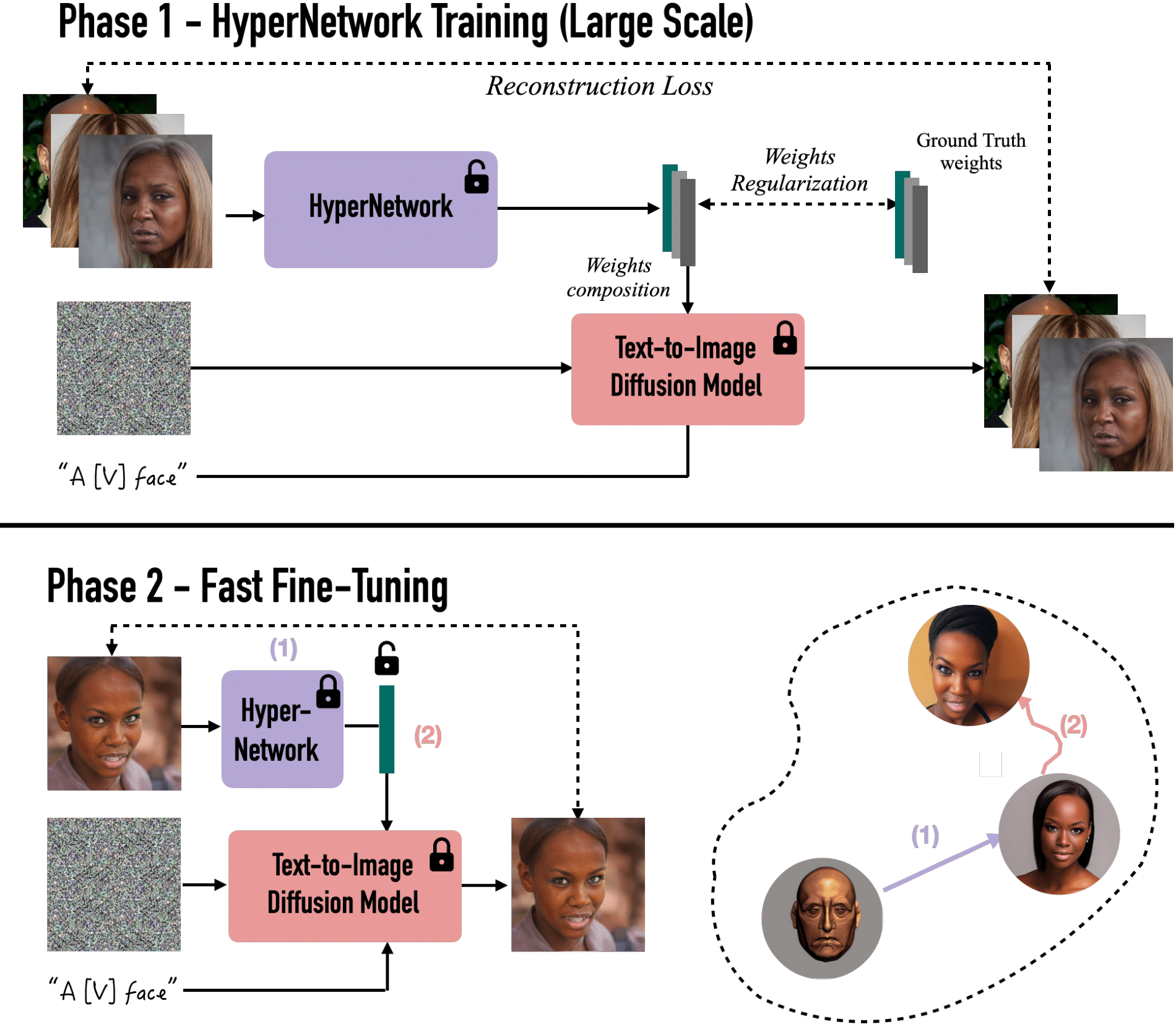} 
    \caption{\textbf{HyperDreamBooth Training and Fast Fine-Tuning.} Phase-1: Training a hypernetwork to predict network weights from a face image, such that a text-to-image diffusion network outputs the person's face from the sentence \textit{"a [v] face"} if the predicted weights are applied to it. We use pre-computed personalized weights for supervision, using an L2 loss, as well as the vanilla diffusion reconstruction loss. Phase-2: Given a face image, our hypernetwork predicts an initial guess for the network weights, which are then fine-tuned using the reconstruction loss to enhance fidelity.}
    \label{fig:hypernet_training}
    \vspace{-0.15in}
\end{figure}
\section{Related Work}
\label{sec:related}

\begin{figure*}[t]
    \centering
    \includegraphics[width=0.87\textwidth]{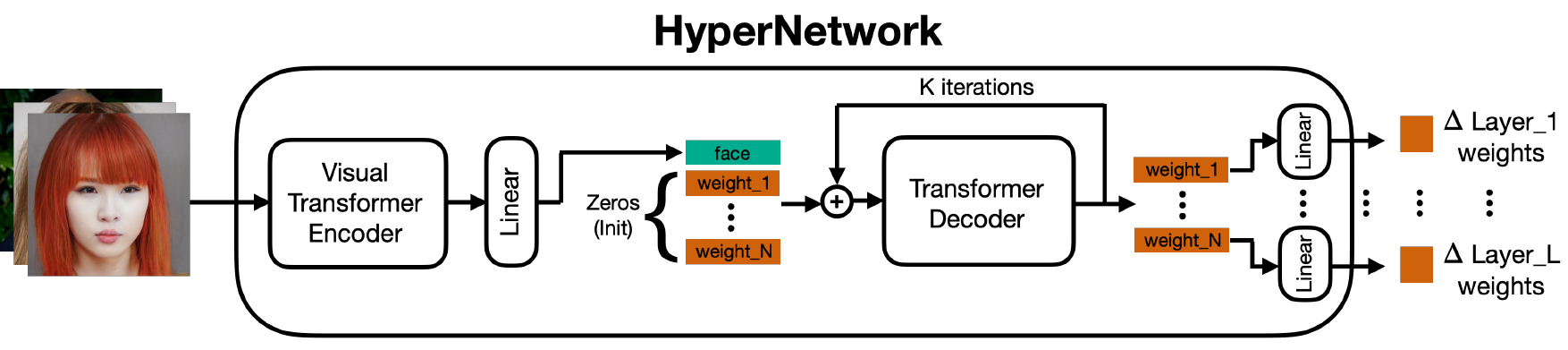}
        \caption{\textbf{HyperNetwork Architecture:} Our hypernetwork consists of a Visual Transformer (ViT) encoder that translates face images into latent face features that are then concatenated to latent layer weight features that are initiated by zeros. A Transformer Decoder receives the sequence of the concatenated features and predicts the values of the weight features in an iterative manner by refining the initial weights with delta predictions. The final layer weight deltas that will be added to the diffusion network are obtained by passing the decoder outputs through learnable linear layers.
        }
    \label{fig:hyper_scheme}
    \vspace{-0.1in}
\end{figure*}


\paragraph{Text-to-Image Models} Several recent models such as Imagen~\cite{saharia2022photorealistic}, DALL-E2~\cite{ramesh2022hierarchical}, Stable Diffusion (SD)~\cite{rombach2022high}, Muse~\cite{chang2023muse}, Parti~\cite{yu2022scaling}, etc., demonstrate excellent image generation capabilities given a text prompt. Some Text-to-Image (T2I) models like SD and Muse also allow conditioning the generation with a given image via an encoder network. Techniques such as ControlNet~\cite{zhang2023adding} propose ways to incorporate new input conditioning such as depth. However, current text and image-based conditioning in these models do not capture sufficient subject details. For ease of experimentation, we demonstrate our HyperDreamBooth on the SD model, given its relatively small size. Yet, the proposed technique is generic and applicable to any T2I model.



\paragraph{Personalization of Generative Models}
Personalized generation aims to create varied images of a specific subject from one or a few reference images. Earlier approaches utilized GANs to manipulate subject images into new contexts. Pivotal tuning~\cite{roich2022pivotal} fine-tunes GANs with inverted latent codes, while~\cite{nitzan2022mystyle} fine-tunes StyleGAN with around 100 images for a personalized generative prior. Casanova et al.~\cite{casanova2021instance} condition a GAN with an input image to produce variations. However, these GAN-based techniques often lack subject fidelity or diverse context in generated images.

HyperNetworks, introduced as auxiliary networks predicting weights for neural networks~\cite{ha2016hypernetworks}, have been applied in image generation tasks akin to personalization, such as StyleGAN inversion~\cite{alaluf2022hyperstyle}, resembling methods that aim to invert an image's latent code for editing in GAN spaces~\cite{alaluf2021restyle}. They have also been used in other tasks such as language modeling~\cite{phang2023hypertuning,ivison2022hint,mu2023learning}.


\paragraph{T2I Personalization via Finetuning}
Recent techniques enhance T2I models for improved subject fidelity and versatile text-based recontextualization. Textual Inversion~\cite{gal2022image} optimizes text embeddings on subject images for image generation, while~\cite{voynov2023p+} explores a richer inversion space capturing more subject details. DreamBooth~\cite{ruiz2022dreambooth} adapts entire network weights for subject fidelity. Various methods, like CustomDiffusion~\cite{kumari2023multi}, SVDiff~\cite{han2023svdiff}, LoRa~\cite{db_lora,hu2021lora}, StyleDrop~\cite{sohn2023styledrop}, and DreamArtist~\cite{dong2022dreamartist}, optimize specific network parts or use specialized tuning strategies. Despite their effectiveness, most of these techniques are slow, taking several minutes per subject for high-quality results.

\paragraph{Fast T2I Personalization}
Several recent and concurrent works aim for faster T2I model personalization. Some, like E4T~\cite{gal2023designing} and ELITE~\cite{wei2023elite}, involve encoder learning followed by complete network finetuning, while our hypernetwork directly predicts low-rank network residuals. SuTI~\cite{chen2023subject} creates a dataset for training a separate network to generate personalized images, but lacks high subject fidelity and affects the original model's integrity. Concurrent work InstantBooth~\cite{shi2023instantbooth} and Taming Encoder~\cite{jia2023taming} introduce conditioning branches for diffusion models, requiring training on large datasets. FastComposer~\cite{xiao2023fastcomposer} focuses on identity blending in multi-subject generation using image encoders. Techniques like~\cite{bansal2023universal}, Face0~\cite{valevski2023face0}, and Celeb-basis~\cite{yuan2023inserting} explore different conditioning or guidance approaches for efficient T2I personalization. However, balancing diversity, fidelity, and adherence to image distribution remains challenging. Our proposed hypernetwork-based approach directly predicts low-rank network residuals for subject-specific adaptation, differing from existing techniques.

\section{Preliminaries}
\label{sec:prelim}

\textbf{Latent Diffusion Models (LDM)}. Text-to-Image (T2I) diffusion models $\mathcal{D}_\theta(\epsilon, \mathbf{c})$ iteratively denoises a
given noise map $\epsilon \in \mathbb{R}^{h\times w}$ into an image $I$ following the description of a text prompt $T$,
which is converted into an input text embedding $\mathbf{c} = \Theta(T)$ using a text encoder $\Theta$.
In this work, we use Stable Diffusion~\cite{rombach2022high}, a specific instatiation of LDM~\cite{rombach2022high}. Briefly, LDM consists of 3 main components: An image encoder that encodes a given image into latent code; a decoder that decodes the latent code back to image pixels; and a U-Net denoising network $\mathcal{D}$ that iteratively denoises a noisy latent code.
See~\cite{rombach2022high} for more details.

\textbf{DreamBooth}~\cite{ruiz2022dreambooth} provides a network fine-tuning strategy to adapt a given T2I denoising network $\mathcal{D}_\theta$ to generate images of a specific subject. At a high-level, DreamBooth optimizes all the diffusion network weights $\theta$ on a few given subject images while also retaining the generalization ability of the original model with class-specific prior preservation loss~\cite{ruiz2022dreambooth}. In the case of Stable Diffusion~\cite{rombach2022high}, this amounts to finetuning the entire denoising UNet has over 1GB of parameters. In addition, DreamBooth on a single subject takes about 5 minutes with 1K training iterations.

\textbf{Low Rank Adaptation (LoRA)}~\cite{hu2021lora,db_lora} provides a memory-efficient and faster technique for DreamBooth.
Specifically, LoRa proposes to finetune the network weight residuals instead of the entire weights. That is, for a layer $l$ with weight matrix $W \in \mathbb{R}^{n \times m}$, LoRa proposes to finetune the residuals $\Delta W$. For diffusion models, LoRa is usually applied for the cross and self-attention layers of the network~\cite{db_lora}.
A key aspect of LoRa is the decomposition of $\Delta W$ matrix into low-rank matrices $A \in \mathbb{R}^{n \times r}$ and $B \in \mathbb{R}^{r \times m}$: $\Delta W = A B$.
The key idea here is that $r << n$ and the combined number of weights in both $A$ and $B$ is much lower than the number of parameters in the original residual $\Delta W$. 
Priors work show that this low-rank residual finetuning is an effective technique that preserves several favorable properties of the original DreamBooth while also being memory-efficient as well as fast, remarkably even when we set $r=1$.
For stable diffusion 1.5 model, LoRA-DreamBooth with $r=1$ has approximately 386K parameters corresponding to only about 1.6MB in size.

\section{Method}
\label{sec:method}

\begin{figure}[t]
    \centering
    \includegraphics[width=\columnwidth]{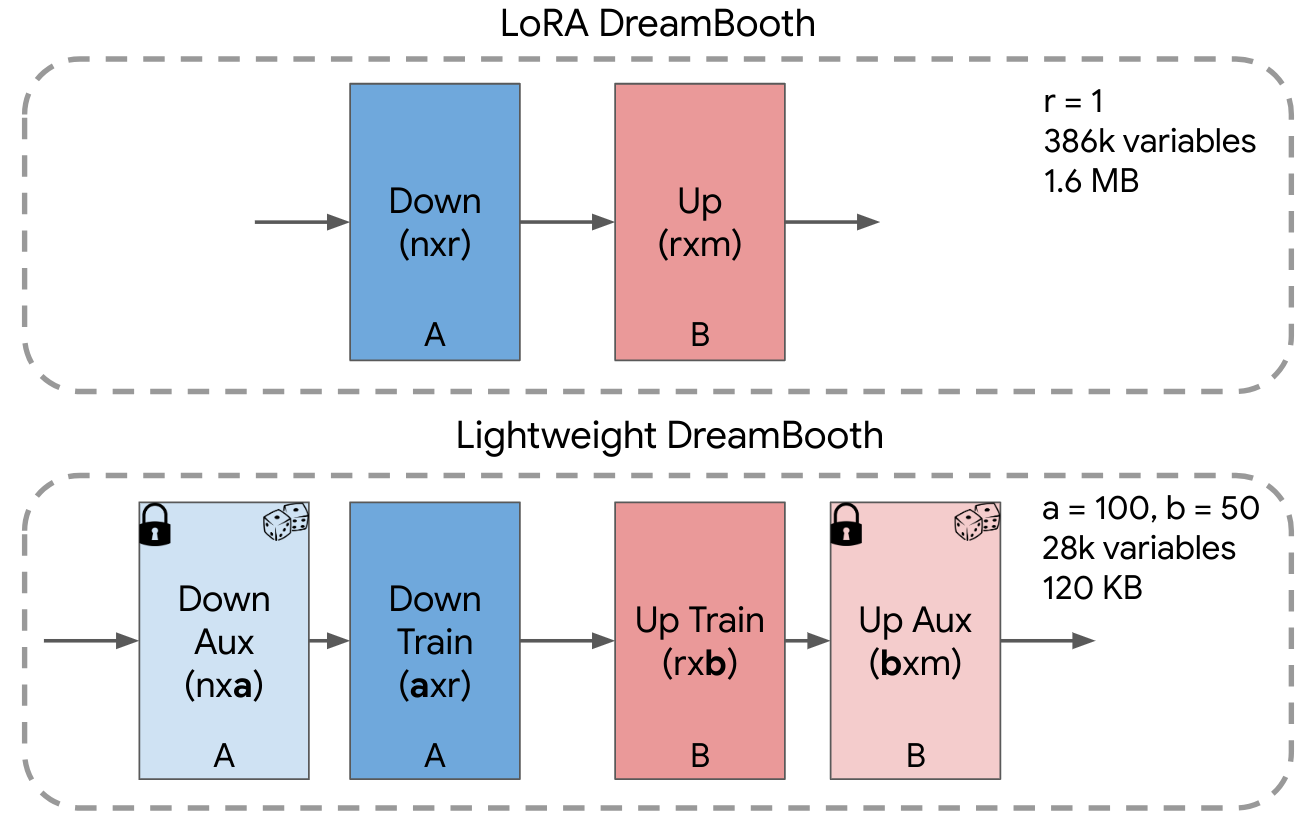}
    \caption{\textbf{Lightweight DreamBooth:} we propose a new low-dimensional weight-space for model personalization generated by a random orthogonal incomplete basis inside LoRA weight-space. This achieves models of roughly 100KB of size (\textbf{0.01\%} of original DreamBooth and \textbf{7.5\%} of LoRA DreamBooth size) and, surprisingly, is sufficient to achieve strong personalization results with solid editability.
    }
    \label{fig:lightweight_dreambooth}
\end{figure}

Our approach consists of 3 core elements which we explain in this section.
We begin by introducing the concept of the Lightweight DreamBooth (LiDB) and demonstrate how the Low-Rank decomposition (LoRa) of the weights can be further decomposed to effectively minimize the number of personalized weights within the model.
Next, we discuss the HyperNetwork training and the architecture the model entails, which enables us to predict the LiDB weights from a single image.
Lastly, we present the concept of rank-relaxed fast fine-tuning, a technique that enables us to significantly amplify the fidelity of the output subject within a few seconds. 
Fig.~\ref{fig:hypernet_training} shows the overview of hypernetwork training followed by fast fine-tuning strategy in our HyperDreamBooth technique.  


\subsection{Lightweight DreamBooth (LiDB)}

\begin{figure*}[t]
    \centering
    \includegraphics[width=0.76\textwidth]{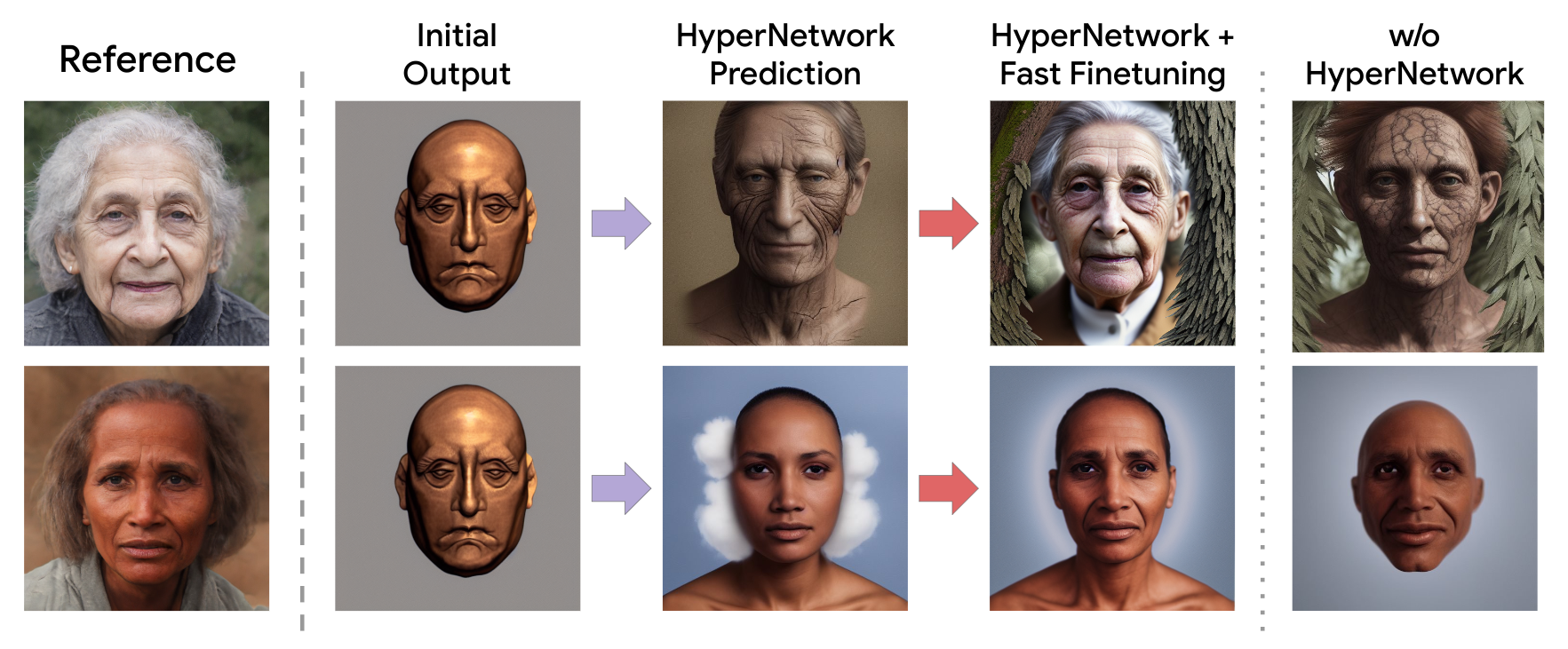}
    \caption{\textbf{HyperNetwork + Fast Finetuning} achieves strong results. Each row displays outputs from initial HyperNetwork prediction (HyperNetwork Prediction column) and after HyperNetwork prediction with fast finetuning (HyperNetwork + Fast Finetuning). Results without the HyperNetwork component highlight its importance.}
    \label{fig:intermediate_hypernet}
    \vspace{-0.1in}
\end{figure*}

Given our objective of generating the personalized subset of weights directly using a HyperNetwork, it would be beneficial to reduce their number to a minimum while maintaining strong results for subject fidelity, editability and style diversity. To this end, we propose a new low-dimensional weight space for model personalization which allows for personalized diffusion models that are 10,000 times smaller than a DreamBooth model and more than 10 times smaller than a LoRA DreamBooth model. Our final version has only 30K variables and takes up only 120 KB of storage space. 

The core idea behind Lightweight DreamBooth (LiDB) is to further decompose the weight-space of a rank-1 LoRa residuals. 
Specifically, we do this using a random orthogonal incomplete basis within the rank-1 LoRA weight-space. We illustrate the idea in Figure~\ref{fig:lightweight_dreambooth}. The approach can also be understood as further decomposing the Down ($A$) and Up ($B$) matrices of LoRA into two matrices each: $A = A_\text{aux}A_\text{train}$ with $A_\text{aux} \in \mathbb{R}^{n\times a}$ and $A_\text{train} \in \mathbb{R}^{a \times r}$ and $B = B_\text{train}B_\text{aux}$ with $B_\text{train} \in \mathbb{R}^{r\times b}$ and $B_\text{aux} \in \mathbb{R}^{b \times m}$.
where the \textit{aux} layers are randomly initialized with row-wise orthogonal vectors and are frozen; and the train layers are learned. Two new hyperparameters are introduced: $a$ and $b$, which we set experimentally.
Thus the weight-residual in a LiDB linear layer is represented as:
\begin{equation}
\Delta W x = A_\text{aux} A_\text{train} B_\text{train} B_\text{aux},
\end{equation}
%
where $r << \text{min}(n, m)$, $a < n$ and $b < m$. $A_\text{aux}$ and $B_\text{aux}$ are randomly initialized with orthogonal row vectors with constant magnitude - and frozen, and $B_\text{train}$ and $A_\text{train}$ are learnable.
Surprisingly, we find that with $a=100$ and $b=50$, which yields models that have only 30K trainable variables and are 120 KB in size, personalization results are strong and maintain subject fidelity, editability and style diversity. We show results for personalization using LiDB in the experiments section.

\subsection{HyperNetwork for Fast Personalization of Text-to-Image Models}

We propose a HyperNetwork for fast personalization of a pre-trained T2I model. Let $\tilde{\theta}$ denote the set of all LiDB 
residual matrices: $A_\text{train}$ and $B_\text{train}$ for each of the cross-attention and self-attention layers of the T2I 
model. In essence, the HyperNetwork $\mathcal{H}_\eta$ with parameters $\eta$ takes the given image $\mathbf{x}$ as input and predicts the LiDB low-rank residuals $\hat{\theta} = \mathcal{H}_\eta(\mathbf{x})$.
The HyperNetwork is trained on a dataset of domain-specific images with a vanilla diffusion denoising loss and a weight-space loss:
\begin{equation}
L(\mathbf{x}) = \alpha ||\mathcal{D}(\mathbf{x} + \epsilon, \mathbf{c}) - \mathbf{x}||_{2}^{2 }+ \beta ||\hat{\theta} - \theta||_2^2,
\end{equation}
where $\mathbf{x}$ is the reference image, $\theta$ are the pre-optimized weight parameters of the personalized model for image $\mathbf{x}$, $\mathcal{D}$ is the diffusion model conditioned on the noisy image $\mathbf{x} + \epsilon$ and the supervisory text-prompt $\mathbf{c}$, and finally $\alpha$ and $\beta$ are hyperparameters that control for the relative weight of each loss. Fig.~\ref{fig:hypernet_training} (top) illustrates the hypernetwork training.

\paragraph{Supervisory Text Prompt} We propose to eschew any type of learned token embedding for this task, and our hypernetwork acts solely to predict the LiDB weights of the diffusion model. We simply propose to condition the learning process ``a [V] face'' for all samples, where [V] is a rare identifier described in \cite{ruiz2022dreambooth}. At inference time variations of this prompt can be used, to insert semantic modifications, for example ``a [V] face in impressionist style''.

\paragraph{HyperNetwork Architecture}
Concretely, as illustrated in Fig.~\ref{fig:hyper_scheme}, we separate the HyperNetwork architecture into two parts: a ViT image encoder and a transformer decoder. We use a ViT-H for the encoder architecture and a 2-hidden layer transformer decoder for the decoder architecture. The transformer decoder is a strong fit for this type of weight prediction task, since the output of a diffusion UNet or Text Encoder is sequentially dependent on the weights of the layers, thus in order to personalize a model there is interdependence of the weights from different layers. In previous work~\cite{ha2016hypernetworks,alaluf2022hyperstyle}, this dependency is not rigorously modeled in the HyperNetwork, whereas with a transformer decoder with a positional embedding, this positional dependency is modeled - similar to dependencies between words in a language model transformer. To the best of our knowledge this is the first use of a transformer decoder as a HyperNetwork.

\paragraph{Iterative Prediction} We find that the HyperNetwork achieves better and more confident predictions given an iterative learning and prediction scenario~\cite{alaluf2022hyperstyle}, where intermediate weight predictions are fed to the HyperNetwork and the network's task is to improve that initial prediction. 
We only perform the image encoding once, and these extracted features $\mathbf{f}$ are then used for all rounds of iterative prediction for the HyperNetwork decoding transformer $\mathcal{T}$. This speeds up training and inference, and we find that it does not affect the quality of results. Specifically, the forward pass of $\mathcal{T}$ becomes:
\begin{equation}
\hat{\theta}_k = \mathcal{T}(\mathbf{f}, \hat{\theta}_{k-1}),
\end{equation}
where $k$ is the current iteration of weight prediction, and terminates once $k=s$, where $s$ is a hyperparameter controlling the maximum amount of iterations. Weights $\theta$ are initialized to zero for $k=0$. Trainable linear layers are used to convert the decoder outputs into the final layer weights. We use the CelebAHQ dataset~\cite{karras2017progressive} for training the HyperNetwork, and find that we only need 15K identities to achieve strong results, much less data than other concurrent methods. For example 100k identities for E4T~\cite{gal2023designing} and 1.43 million identities for InstantBooth~\cite{shi2023instantbooth}.

\subsection{Rank-Relaxed Fast Finetuning}
We find that the initial HyperNetwork prediction is in great measure directionally correct and generates faces with similar semantic attributes (gender, facial hair, hair color, skin color, etc.) as the target face consistently. Nevertheless, fine details are not sufficiently captured. We propose a final fast finetuning step in order to capture such details, which is magnitudes faster than DreamBooth, but achieves virtually identical results with strong subject fidelity, editability and style diversity. 
Specifically, we first predict personalized diffusion model weights $\hat{\theta} = \mathcal{H}(\mathbf{x})$ and then subsequently finetune the weights using the diffusion denoising loss $L(\mathbf{x}) = ||\mathcal{D}_{\hat{\theta}}(\mathbf{x} + \epsilon, \mathbf{c}) - \mathbf{x}||_2^2$. A key contribution of our work is the idea of \textit{rank-relaxed} finetuning, where we relax the rank of the LoRA model from $r=1$ to $r>1$ before fast finetuning. Specifically, we add the predicted HyperNetwork weights to the overall weights of the model, and then perform LoRA finetuning with a new higher rank. This expands the capability of our method of approximating high-frequency details of the subject, giving higher subject fidelity than methods that are locked to lower ranks of weight updates. To the best of our knowledge we are the first to propose such rank-relaxed LoRA models.

We use the same supervision text prompt ``a [V] face'' this fast finetuning step. We find that given the HyperNetwork initialization, fast finetuning can be done in 40 iterations, which is \textbf{25x} faster than DreamBooth~\cite{ruiz2022dreambooth} and LoRA DreamBooth~\cite{db_lora}. We show an example of initial, intermediate and final results in Figure~\ref{fig:intermediate_hypernet}.
\vspace{-0.05in}
\section{Experiments}
\label{sec:exp}
\vspace{-0.05in}

\begin{figure*}[t]
    \centering
    \includegraphics[width=0.68\textwidth]{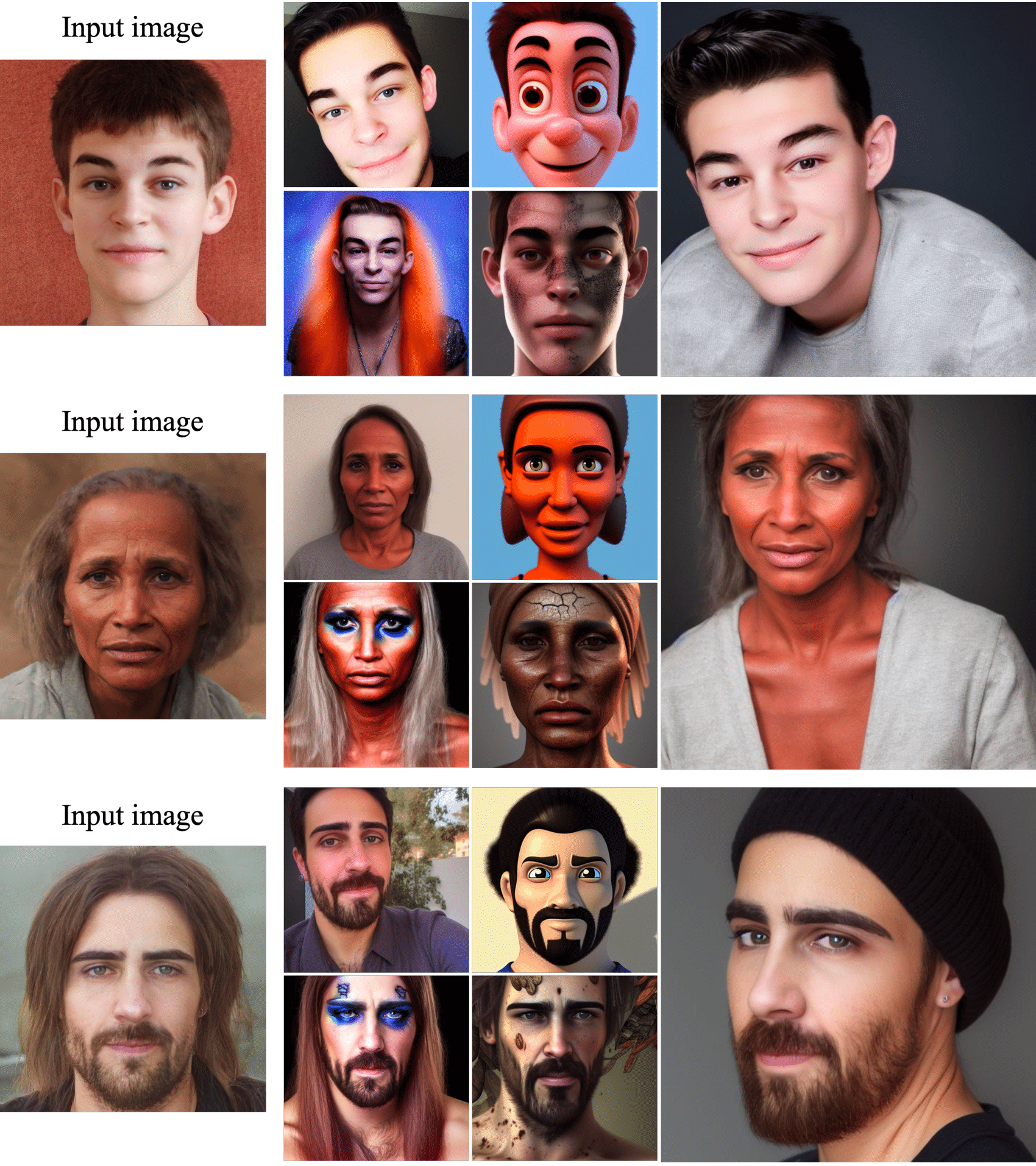}
    \caption{\textbf{Results Gallery:} Our method can generate novel artistic and stylized results of diverse subjects (depicted in an input image, left) with considerable editability while maintaining the integrity to the subject's key facial characteristics. The output images were generated with the following captions (top-left to bottom-right): \textit{``An Instagram selfie of a [V] face", ``A Pixar character of a [V] face", ``A [V] face with bark skin", ``A [V] face as a rock star"}. Rightmost: \textit{``A professional shot of a [V] face".}
    }
    \label{fig:results}
    \vspace{-0.1in}
\end{figure*}

We implement our HyperDreamBooth on the Stable Diffusion v1.5 diffusion model and we predict the LoRa weights for all cross and self-attention layers of the diffusion UNet as well as the CLIP text encoder. For privacy reasons, all face images used for visuals are synthetic, from the SFHQ dataset~\cite{david_beniaguev_2022_SFHQ}. For training, we use 15K images from CelebA-HQ~\cite{karras2017progressive}.

\subsection{Subject Personalization Results}
Our method achieves strong personalization results for widely diverse faces, with performance that is identically or surpasses that of the state-of-the art optimization driven methods~\cite{ruiz2022dreambooth, gal2022image, gal2023designing}. Moreover, we achieve very strong editability, with semantic transformations of face identities into highly different domains such as figurines and animated characters, and we conserve the strong style prior of the model which allows for a wide variety of style generations. We show results in Figure~\ref{fig:results}.


\begin{figure*}[t]
    \centering
    \includegraphics[width=0.8\textwidth]{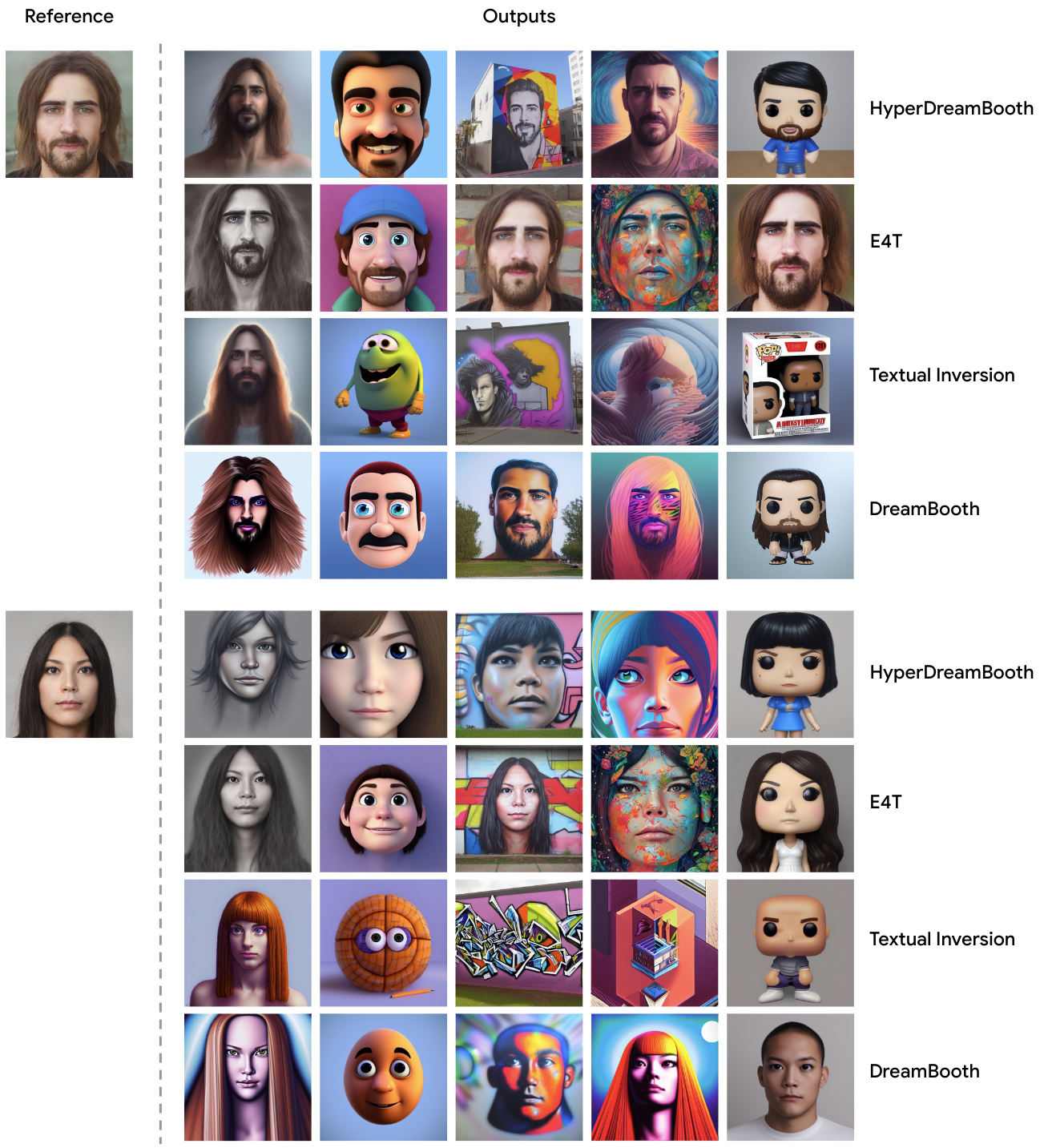}
    \caption{\textbf{Qualitative Comparison:} We compare random samples from our method (HyperDreamBooth), E4T~\cite{gal2023designing}, DreamBooth~\cite{ruiz2022dreambooth} and Textual Inversion~\cite{gal2022image} for two different identities and five different stylistic prompts. We observe that our method generally achieves very strong editability while preserving identity, generally surpassing competing methods in the single-reference regime. E4T shows strong performance but can tend to overfit to the reference head pose and realistic appearance, even when the image should be strongly stylized.
    } 
    \vspace{-0.1in}
    \label{fig:comparison}
\end{figure*}

\begin{table}[t]
\caption{\textbf{Comparisons.} We compare our method for face identity preservation (Face Rec.), subject fidelity (DINO, CLIP-I) and prompt fidelity (CLIP-T) to DreamBooth and Textual Inversion. We find that our method preserves identity and subject fidelity more closely, while achieving a higher score in prompt fidelity.
\label{table:main_comparison_experiment}}
\centering
\resizebox{\columnwidth}{!}{
  \begin{tabular}{lccccc}
    \toprule
    Method & Face Rec. $\uparrow$ & DINO $\uparrow$ & CLIP-I $\uparrow$ & CLIP-T $\uparrow$\\
    \midrule
    Ours & \textbf{0.655} & \textbf{0.473} & \textbf{0.577} & \textbf{0.286} \\
    DreamBooth & 0.618 & 0.441 & 0.546 & 0.282 \\
    Textual Inversion & 0.623 & 0.289 & 0.472 & 0.277 \\
    \bottomrule
  \end{tabular}
}
\end{table}

\begin{table}[t]
\caption{\textbf{Comparisons with DreamBooth}. We compare our method to differently tuned versions of DreamBooth that minimize optimization time. Altering hyperparameters by increasing the learning rate and decreasing iterations leads to degraded results in DreamBooth. DreamBooth-Agg-1 uses 400 iterations and DreamBooth-Agg-2 uses 40 iterations as opposed to the normal 1200 iterations used in our vanilla DreamBooth.
\label{table:comparison_different_training}}

\centering
\resizebox{\columnwidth}{!}{
  \begin{tabular}{lccccc}
    \toprule
    Method & Face Rec. $\uparrow$ & DINO $\uparrow$ & CLIP-I $\uparrow$ & CLIP-T $\uparrow$\\
    \midrule
    Ours & \textbf{0.655} & \textbf{0.473} & \textbf{0.577} & 0.286 \\
    DreamBooth & 0.618 & 0.441 & 0.546 & 0.282 \\
    DreamBooth-Agg-1 & 0.615 & 0.323 & 0.431 & \textbf{0.313} \\
    DreamBooth-Agg-2 & 0.616 & 0.360 & 0.467 & 0.302 \\
    \bottomrule
  \end{tabular}
}
\end{table}

\begin{table}[t]
\caption{{\textbf{HyperNetwork Ablation}. We ablate components: No Hyper (without hypernetwork at test-time), Only Hyper (using hypernetwork prediction without fast finetuning), and our full method without iterative prediction (k=1). Our full method performs best for all fidelity metrics, with No Hyper achieving slightly better prompt following.}
\label{table:hypernetwork_ablation}}
\centering
\resizebox{0.9\columnwidth}{!}{
  \begin{tabular}{lccccc}
    \toprule
    Method & Face Rec. $\uparrow$ & DINO $\uparrow$ & CLIP-I $\uparrow$ & CLIP-T $\uparrow$\\
    \midrule
    Ours & \textbf{0.655} & \textbf{0.473} & \textbf{0.577} & 0.286 \\
    No Hyper & 0.647 & 0.392 & 0.498 & \textbf{0.299} \\
    Only Hyper & 0.631 & 0.414 & 0.501 & 0.298 \\
    Ours (k=1) & 0.648 & 0.464 & 0.570 & 0.288 \\
    \bottomrule
  \end{tabular}
}
\end{table}

\begin{table}[b]
\caption{\textbf{User Study}. Given limitations of face recognition networks (stylized faces are OOD), we conduct an identity fidelity user study comparing our stylized generations against DB and TI. Our approach generally receives higher user preference.
\label{table:user_study}}

\centering
\resizebox{\columnwidth}{!}{
  \begin{tabular}{lccc|ccc}
    \toprule
     &  Ours  &  DB  &  Undecided  &  Ours  &  TI  &  Undecided \\
    \midrule
    Pref. $\uparrow$ & \textbf{64.8\%} & 23.3\% & 11.9\% & \textbf{70.6\%} & 21.6\% & 7.8\% \\
    \bottomrule
  \end{tabular}
}
\end{table}

\begin{table}[b]
\caption{\textbf{User Stylization and Identity Preference}. We compare the user preference of stylization and identity between our approach and the SoTA approach E4T. Users generally prefer our method.
\label{table:user_study_e4t}}
\centering
\resizebox{0.75\columnwidth}{!}{
  \begin{tabular}{lccc}
    \toprule
     & Ours & E4T & Undecided \\
    \midrule
    Preference $\uparrow$ &  \textbf{60.0\%} & 37.5\% & 2.5\% \\
    \bottomrule
  \end{tabular}
}
\end{table}


\subsection{Comparisons}
\paragraph{Qualitative Comparisons}
We compare our method to Textual Inversion~\cite{gal2022image}, DreamBooth~\cite{ruiz2022dreambooth} and E4T~\cite{gal2023designing}. Results are shown in Figure~\ref{fig:comparison}. We observe that our method strongly outperforms both Textual Inversion and DreamBooth generally, in the one-input-image regime - and obtains strong results compared to E4T, especially in cases where E4T overfits to the reference face pose and realistic appearance, even though the output should be highly stylized.

\paragraph{Quantitative Comparisons and Ablations}
We compare our method to Textual Inversion and DreamBooth using face recognition metrics (``Face Rec.'' from a VGGFace2 Inception ResNet), along with DINO, CLIP-I, and CLIP-T metrics \cite{ruiz2022dreambooth}. Using 100 CelebAHQ identities and 30 prompts (style modification and recontextualization), totaling 30,000 samples, Table~\ref{table:main_comparison_experiment} illustrates our approach outperforming in all metrics. However, face recognition metrics are relatively weak here due to network training limitations (realistic face bias). To compensate, we conduct a user study (details below).

We also conduct comparisons with more aggressive DreamBooth training with altered iterations and learning rates. Specifically, DreamBooth-Agg-1 (400 iterations) and DreamBooth-Agg-2 (40 iterations) differ from our 1200-iteration vanilla DreamBooth. Table~\ref{table:comparison_different_training} reveals that aggressive DreamBooth training without our HyperNetwork initialization generally degrades results.

Additionally, we show an ablation study that explores our method's components: removing the HyperNetwork (No Hyper), utilizing only the HyperNetwork without finetuning (Only Hyper), and our full setup without iterative predictions (k=1). Table~\ref{table:hypernetwork_ablation} demonstrates that our complete setup achieves superior subject fidelity, albeit with a slightly lower prompt following metric.

\paragraph{User Study}
We conduct a user study for face identity preservation of outputs and compare our method to DreamBooth and Textual Inversion. Specifically, we present the reference face image and two random generations using the same prompt from our method and the baseline, and ask the user to rate which one has most similar face identity to the reference face image. We test a total of 25 identities, and query 5 users per question, with a total of 1,000 sample pairs evaluated. We take the majority vote for each pair. We present our results in Table~\ref{table:user_study}, where we show a strong preference for face identity preservation of our method.

Finally, we present a user study for overall preference of both subject fidelity and style fidelity and compare our approach to the published state-of-the-art E4T method~\cite{gal2023designing} on a set of identities from the SFHQ dataset, with E4T kindly run by the authors. We present both the reference subject image as well as a reference style image and ask users which output they prefer with respect to both identity preservation and style preservation. We test 10 identities, 4 prompts per identity, and query 15 users per question, totaling 600 samples. Results are shown in Table~\ref{table:user_study_e4t}, where we observe a preference of users for our method. Although E4T is a method that achieves strong results and preserves identity well, we observe slightly less qualitative editability as well as some consistency errors with hard prompts. Note our method is trained on 15k identities vs.~100k identities for E4T.


\vspace{-0.05in}
\section{Conclusion}
\label{sec:conclusion}
\vspace{-0.05in}

In this work, we presented \textit{HyperDreamBooth} a new method for fast and lightweight subject personalization of diffusion models. It leverages a HyperNetwork to generate Lightweight DreamBooth (LiDB) parameters for a diffusion model with a subsequent fast rank-relaxed finetuning that achieves a sharp reduction in size and speed compared to DreamBooth and other optimization-based personalization work. We showed that it produces high-quality and diverse images of faces with different styles and semantic modifications, while preserving subject details and model integrity. 
{
    \small
    \bibliographystyle{ieeenat_fullname}
    \bibliography{main}

\begin{thebibliography}{35}
\providecommand{\natexlab}[1]{#1}
\providecommand{\url}[1]{\texttt{#1}}
\expandafter\ifx\csname urlstyle\endcsname\relax
  \providecommand{\doi}[1]{doi: #1}\else
  \providecommand{\doi}{doi: \begingroup \urlstyle{rm}\Url}\fi

\bibitem[db_(2022)]{db_lora}
Low-rank adaptation for fast text-to-image diffusion fine-tuning.
\newblock \url{https://github.com/cloneofsimo/lora}, 2022.

\bibitem[Alaluf et~al.(2021)Alaluf, Patashnik, and Cohen-Or]{alaluf2021restyle}
Yuval Alaluf, Or Patashnik, and Daniel Cohen-Or.
\newblock Restyle: A residual-based stylegan encoder via iterative refinement.
\newblock In \emph{Proceedings of the IEEE/CVF International Conference on Computer Vision}, pages 6711--6720, 2021.

\bibitem[Alaluf et~al.(2022)Alaluf, Tov, Mokady, Gal, and Bermano]{alaluf2022hyperstyle}
Yuval Alaluf, Omer Tov, Ron Mokady, Rinon Gal, and Amit Bermano.
\newblock Hyperstyle: Stylegan inversion with hypernetworks for real image editing.
\newblock In \emph{Proceedings of the IEEE/CVF conference on computer Vision and pattern recognition}, pages 18511--18521, 2022.

\bibitem[Bansal et~al.(2023)Bansal, Chu, Schwarzschild, Sengupta, Goldblum, Geiping, and Goldstein]{bansal2023universal}
Arpit Bansal, Hong-Min Chu, Avi Schwarzschild, Soumyadip Sengupta, Micah Goldblum, Jonas Geiping, and Tom Goldstein.
\newblock Universal guidance for diffusion models.
\newblock In \emph{Proceedings of the IEEE/CVF Conference on Computer Vision and Pattern Recognition}, pages 843--852, 2023.

\bibitem[Beniaguev(2022)]{david_beniaguev_2022_SFHQ}
David Beniaguev.
\newblock Synthetic faces high quality (sfhq) dataset.
\newblock \url{https://github.com/SelfishGene/SFHQ-dataset}, 2022.

\bibitem[Casanova et~al.(2021)Casanova, Careil, Verbeek, Drozdzal, and Romero~Soriano]{casanova2021instance}
Arantxa Casanova, Marlene Careil, Jakob Verbeek, Michal Drozdzal, and Adriana Romero~Soriano.
\newblock Instance-conditioned gan.
\newblock \emph{Advances in Neural Information Processing Systems}, 34:\penalty0 27517--27529, 2021.

\bibitem[Chang et~al.(2023)Chang, Zhang, Barber, Maschinot, Lezama, Jiang, Yang, Murphy, Freeman, Rubinstein, et~al.]{chang2023muse}
Huiwen Chang, Han Zhang, Jarred Barber, AJ Maschinot, Jose Lezama, Lu Jiang, Ming-Hsuan Yang, Kevin Murphy, William~T Freeman, Michael Rubinstein, et~al.
\newblock Muse: Text-to-image generation via masked generative transformers.
\newblock \emph{arXiv preprint arXiv:2301.00704}, 2023.

\bibitem[Chen et~al.(2023)Chen, Hu, Li, Ruiz, Jia, Chang, and Cohen]{chen2023subject}
Wenhu Chen, Hexiang Hu, Yandong Li, Nataniel Ruiz, Xuhui Jia, Ming-Wei Chang, and William~W Cohen.
\newblock Subject-driven text-to-image generation via apprenticeship learning.
\newblock \emph{arXiv preprint arXiv:2304.00186}, 2023.

\bibitem[Dong et~al.(2022)Dong, Wei, and Lin]{dong2022dreamartist}
Ziyi Dong, Pengxu Wei, and Liang Lin.
\newblock Dreamartist: Towards controllable one-shot text-to-image generation via contrastive prompt-tuning.
\newblock \emph{arXiv preprint arXiv:2211.11337}, 2022.

\bibitem[Gal et~al.(2022)Gal, Alaluf, Atzmon, Patashnik, Bermano, Chechik, and Cohen-Or]{gal2022image}
Rinon Gal, Yuval Alaluf, Yuval Atzmon, Or Patashnik, Amit~H Bermano, Gal Chechik, and Daniel Cohen-Or.
\newblock An image is worth one word: Personalizing text-to-image generation using textual inversion.
\newblock \emph{arXiv preprint arXiv:2208.01618}, 2022.

\bibitem[Gal et~al.(2023)Gal, Arar, Atzmon, Bermano, Chechik, and Cohen-Or]{gal2023designing}
Rinon Gal, Moab Arar, Yuval Atzmon, Amit~H Bermano, Gal Chechik, and Daniel Cohen-Or.
\newblock Designing an encoder for fast personalization of text-to-image models.
\newblock \emph{arXiv preprint arXiv:2302.12228}, 2023.

\bibitem[Ha et~al.(2016)Ha, Dai, and Le]{ha2016hypernetworks}
David Ha, Andrew Dai, and Quoc~V Le.
\newblock Hypernetworks.
\newblock \emph{arXiv preprint arXiv:1609.09106}, 2016.

\bibitem[Han et~al.(2023)Han, Li, Zhang, Milanfar, Metaxas, and Yang]{han2023svdiff}
Ligong Han, Yinxiao Li, Han Zhang, Peyman Milanfar, Dimitris Metaxas, and Feng Yang.
\newblock Svdiff: Compact parameter space for diffusion fine-tuning.
\newblock \emph{arXiv preprint arXiv:2303.11305}, 2023.

\bibitem[Hu et~al.(2021)Hu, Shen, Wallis, Allen-Zhu, Li, Wang, Wang, and Chen]{hu2021lora}
Edward~J Hu, Yelong Shen, Phillip Wallis, Zeyuan Allen-Zhu, Yuanzhi Li, Shean Wang, Lu Wang, and Weizhu Chen.
\newblock Lora: Low-rank adaptation of large language models.
\newblock \emph{arXiv preprint arXiv:2106.09685}, 2021.

\bibitem[Ivison et~al.(2022)Ivison, Bhagia, Wang, Hajishirzi, and Peters]{ivison2022hint}
Hamish Ivison, Akshita Bhagia, Yizhong Wang, Hannaneh Hajishirzi, and Matthew Peters.
\newblock Hint: Hypernetwork instruction tuning for efficient zero-shot generalisation.
\newblock \emph{arXiv preprint arXiv:2212.10315}, 2022.

\bibitem[Jia et~al.(2023)Jia, Zhao, Chan, Li, Zhang, Gong, Hou, Wang, and Su]{jia2023taming}
Xuhui Jia, Yang Zhao, Kelvin~CK Chan, Yandong Li, Han Zhang, Boqing Gong, Tingbo Hou, Huisheng Wang, and Yu-Chuan Su.
\newblock Taming encoder for zero fine-tuning image customization with text-to-image diffusion models.
\newblock \emph{arXiv preprint arXiv:2304.02642}, 2023.

\bibitem[Karras et~al.(2017)Karras, Aila, Laine, and Lehtinen]{karras2017progressive}
Tero Karras, Timo Aila, Samuli Laine, and Jaakko Lehtinen.
\newblock Progressive growing of gans for improved quality, stability, and variation.
\newblock \emph{arXiv preprint arXiv:1710.10196}, 2017.

\bibitem[Kumari et~al.(2023)Kumari, Zhang, Zhang, Shechtman, and Zhu]{kumari2023multi}
Nupur Kumari, Bingliang Zhang, Richard Zhang, Eli Shechtman, and Jun-Yan Zhu.
\newblock Multi-concept customization of text-to-image diffusion.
\newblock In \emph{Proceedings of the IEEE/CVF Conference on Computer Vision and Pattern Recognition}, pages 1931--1941, 2023.

\bibitem[Mu et~al.(2023)Mu, Li, and Goodman]{mu2023learning}
Jesse Mu, Xiang~Lisa Li, and Noah Goodman.
\newblock Learning to compress prompts with gist tokens.
\newblock \emph{arXiv preprint arXiv:2304.08467}, 2023.

\bibitem[Nitzan et~al.(2022)Nitzan, Aberman, He, Liba, Yarom, Gandelsman, Mosseri, Pritch, and Cohen-Or]{nitzan2022mystyle}
Yotam Nitzan, Kfir Aberman, Qiurui He, Orly Liba, Michal Yarom, Yossi Gandelsman, Inbar Mosseri, Yael Pritch, and Daniel Cohen-Or.
\newblock Mystyle: A personalized generative prior.
\newblock \emph{ACM Transactions on Graphics (TOG)}, 41\penalty0 (6):\penalty0 1--10, 2022.

\bibitem[Phang et~al.(2023)Phang, Mao, He, and Chen]{phang2023hypertuning}
Jason Phang, Yi Mao, Pengcheng He, and Weizhu Chen.
\newblock Hypertuning: Toward adapting large language models without back-propagation.
\newblock In \emph{International Conference on Machine Learning}, pages 27854--27875. PMLR, 2023.

\bibitem[Ramesh et~al.(2022)Ramesh, Dhariwal, Nichol, Chu, and Chen]{ramesh2022hierarchical}
Aditya Ramesh, Prafulla Dhariwal, Alex Nichol, Casey Chu, and Mark Chen.
\newblock Hierarchical text-conditional image generation with clip latents.
\newblock \emph{arXiv preprint arXiv:2204.06125}, 2022.

\bibitem[Roich et~al.(2022)Roich, Mokady, Bermano, and Cohen-Or]{roich2022pivotal}
Daniel Roich, Ron Mokady, Amit~H Bermano, and Daniel Cohen-Or.
\newblock Pivotal tuning for latent-based editing of real images.
\newblock \emph{ACM Transactions on graphics (TOG)}, 42\penalty0 (1):\penalty0 1--13, 2022.

\bibitem[Rombach et~al.(2022)Rombach, Blattmann, Lorenz, Esser, and Ommer]{rombach2022high}
Robin Rombach, Andreas Blattmann, Dominik Lorenz, Patrick Esser, and Bj{\"o}rn Ommer.
\newblock High-resolution image synthesis with latent diffusion models.
\newblock In \emph{Proceedings of the IEEE/CVF conference on computer vision and pattern recognition}, pages 10684--10695, 2022.

\bibitem[Ruiz et~al.(2022)Ruiz, Li, Jampani, Pritch, Rubinstein, and Aberman]{ruiz2022dreambooth}
Nataniel Ruiz, Yuanzhen Li, Varun Jampani, Yael Pritch, Michael Rubinstein, and Kfir Aberman.
\newblock Dreambooth: Fine tuning text-to-image diffusion models for subject-driven generation.
\newblock 2022.

\bibitem[Saharia et~al.(2022)Saharia, Chan, Saxena, Li, Whang, Denton, Ghasemipour, Gontijo~Lopes, Karagol~Ayan, Salimans, et~al.]{saharia2022photorealistic}
Chitwan Saharia, William Chan, Saurabh Saxena, Lala Li, Jay Whang, Emily~L Denton, Kamyar Ghasemipour, Raphael Gontijo~Lopes, Burcu Karagol~Ayan, Tim Salimans, et~al.
\newblock Photorealistic text-to-image diffusion models with deep language understanding.
\newblock \emph{Advances in Neural Information Processing Systems}, 35:\penalty0 36479--36494, 2022.

\bibitem[Shi et~al.(2023)Shi, Xiong, Lin, and Jung]{shi2023instantbooth}
Jing Shi, Wei Xiong, Zhe Lin, and Hyun~Joon Jung.
\newblock Instantbooth: Personalized text-to-image generation without test-time finetuning.
\newblock \emph{arXiv preprint arXiv:2304.03411}, 2023.

\bibitem[Sohn et~al.(2023)Sohn, Ruiz, Lee, Chin, Blok, Chang, Barber, Jiang, Entis, Li, Hao, Essa, Rubinstein, and Krishnan]{sohn2023styledrop}
Kihyuk Sohn, Nataniel Ruiz, Kimin Lee, Daniel~Castro Chin, Irina Blok, Huiwen Chang, Jarred Barber, Lu Jiang, Glenn Entis, Yuanzhen Li, Yuan Hao, Irfan Essa, Michael Rubinstein, and Dilip Krishnan.
\newblock Styledrop: Text-to-image generation in any style.
\newblock \emph{arXiv preprint arXiv:2306.00983}, 2023.

\bibitem[Valevski et~al.(2023)Valevski, Wasserman, Matias, and Leviathan]{valevski2023face0}
Dani Valevski, Danny Wasserman, Yossi Matias, and Yaniv Leviathan.
\newblock Face0: Instantaneously conditioning a text-to-image model on a face.
\newblock \emph{arXiv preprint arXiv:2306.06638}, 2023.

\bibitem[Voynov et~al.(2023)Voynov, Chu, Cohen-Or, and Aberman]{voynov2023p+}
Andrey Voynov, Qinghao Chu, Daniel Cohen-Or, and Kfir Aberman.
\newblock $ p+ $: Extended textual conditioning in text-to-image generation.
\newblock \emph{arXiv preprint arXiv:2303.09522}, 2023.

\bibitem[Wei et~al.(2023)Wei, Zhang, Ji, Bai, Zhang, and Zuo]{wei2023elite}
Yuxiang Wei, Yabo Zhang, Zhilong Ji, Jinfeng Bai, Lei Zhang, and Wangmeng Zuo.
\newblock Elite: Encoding visual concepts into textual embeddings for customized text-to-image generation.
\newblock \emph{arXiv preprint arXiv:2302.13848}, 2023.

\bibitem[Xiao et~al.(2023)Xiao, Yin, Freeman, Durand, and Han]{xiao2023fastcomposer}
Guangxuan Xiao, Tianwei Yin, William~T Freeman, Fr{\'e}do Durand, and Song Han.
\newblock Fastcomposer: Tuning-free multi-subject image generation with localized attention.
\newblock \emph{arXiv preprint arXiv:2305.10431}, 2023.

\bibitem[Yu et~al.(2022)Yu, Xu, Koh, Luong, Baid, Wang, Vasudevan, Ku, Yang, Ayan, et~al.]{yu2022scaling}
Jiahui Yu, Yuanzhong Xu, Jing~Yu Koh, Thang Luong, Gunjan Baid, Zirui Wang, Vijay Vasudevan, Alexander Ku, Yinfei Yang, Burcu~Karagol Ayan, et~al.
\newblock Scaling autoregressive models for content-rich text-to-image generation.
\newblock \emph{arXiv preprint arXiv:2206.10789}, 2022.

\bibitem[Yuan et~al.(2023)Yuan, Cun, Zhang, Li, Qi, Wang, Shan, and Zheng]{yuan2023inserting}
Ge Yuan, Xiaodong Cun, Yong Zhang, Maomao Li, Chenyang Qi, Xintao Wang, Ying Shan, and Huicheng Zheng.
\newblock Inserting anybody in diffusion models via celeb basis.
\newblock \emph{arXiv preprint arXiv:2306.00926}, 2023.

\bibitem[Zhang and Agrawala(2023)]{zhang2023adding}
Lvmin Zhang and Maneesh Agrawala.
\newblock Adding conditional control to text-to-image diffusion models.
\newblock \emph{arXiv preprint arXiv:2302.05543}, 2023.

\end{thebibliography}
}


\end{document}


\title{\vspace{-.25in}Supplementary Material for HyperDreamBooth: HyperNetworks \\ for Fast Personalization of Text-to-Image Models\vspace{-.1in}}

\author{
}
\maketitle

\section*{Additional Experiments}
\label{sec:additional_experiments}

\begin{figure*}[H]
    \centering
    \includegraphics[width=0.9\textwidth]{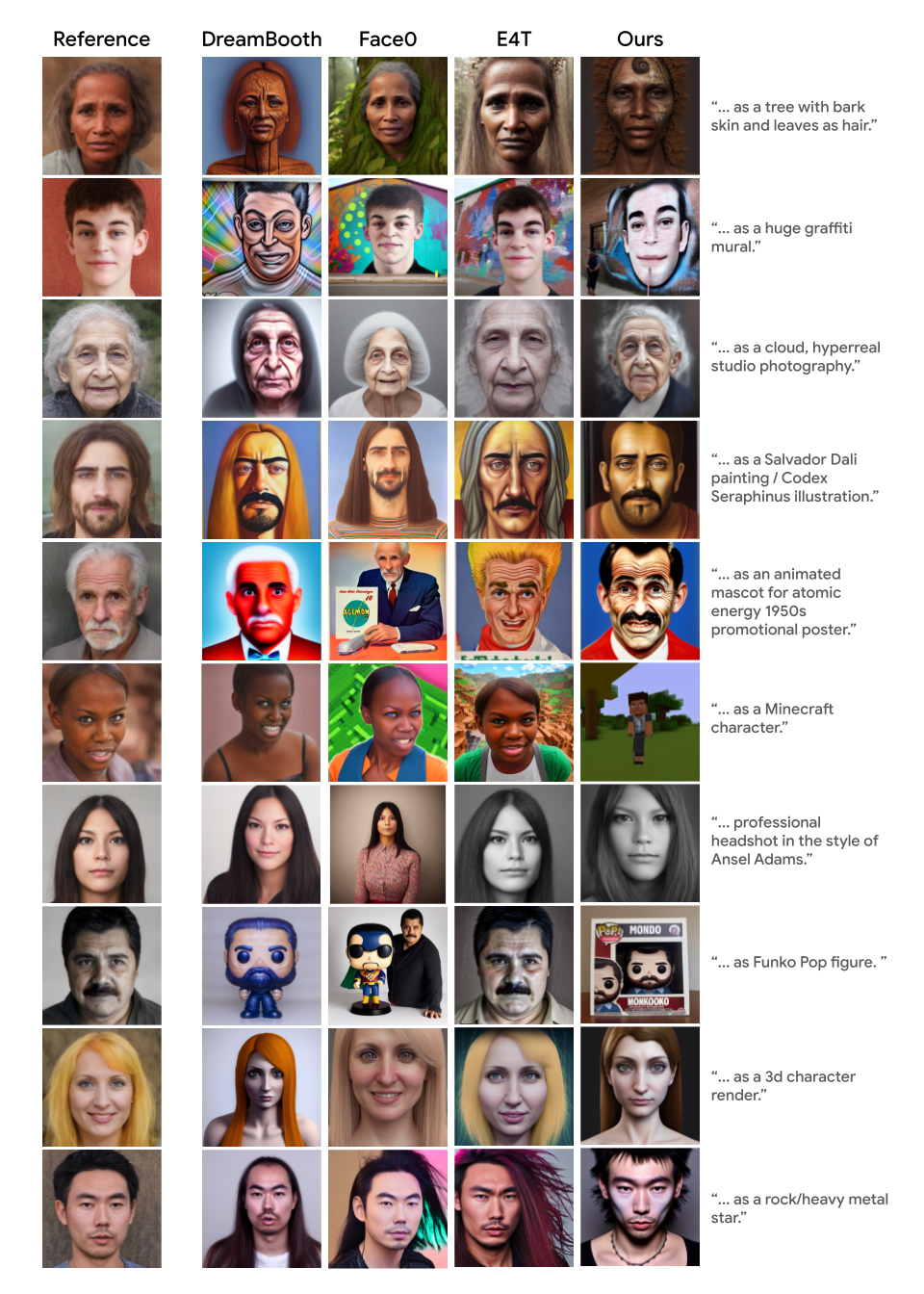}
    \caption{\textbf{Qualitative Comparison:} We compare random samples from DreamBooth~\cite{ruiz2022dreambooth}, Face0 (concurrent work)~\cite{valevski2023face0}, the state-of-the-art published work E4T~\cite{gal2023designing} and our method (HyperDreamBooth).
    } 
    \label{fig:comparison_supp}
\end{figure*}

\paragraph{Qualitative Comparisons}
We provide more extensive qualitative comparisons over 10 identities of SFHQ and different stylistic prompts for our method, E4T~\cite{gal2023designing} and concurrent work Face0~\cite{valevski2023face0}. We show results in Figure~\ref{fig:comparison_supp}. Our method typically demonstrates high editability and maintains identity effectively in the single-reference scenario, outperforming other methods. While E4T and Face0 yield impressive results, they often suffer from overfitting to realistic faces, which restricts their editability. Additionally, these methods, and especially Face0, tend to produce frontal outputs that closely resemble the pose of the reference input image.

\paragraph{Quantitative Comparisons and Ablations}
In this section, we present a quantitative comparison of our method with the E4T method~\cite{gal2023designing} on a subset of 10 identities from the SFHQ dataset, with E4T kindly provided by the authors. We evaluate performance in generating stylistic portraits for 20 style prompts. Our evaluation metrics include face identity preservation (``Face Rec.'' from a VGGFace2 Inception ResNet), subject fidelity (assessed using DINO and CLIP-I metrics), and prompt fidelity (evaluated using CLIP-T). The results of this comparison are detailed in Table~\ref{table:e4t_comparison}. Note that our method is trained on 15k identities vs.~100k identities for E4T. WE note that these metrics can be relatively suboptimal in the stylized portrait scenario, with face recognition metrics overfitting to the realistic domain, and DINO and CLIP-I metrics strongly preferring very similar structure or high-level semantics instead of focusing on specific subject and style details of images. In the main paper we include what we would consider a stronger user study experiment, where users choose the method that they prefer in terms of subject and style fidelity.

Our analysis reveals that our method obtains higher face identity and prompt fidelity metrics compared to E4T. In terms of similarity to the reference subject image, measured by DINO and CLIP-I metrics, our method shows lower scores compared to E4T. This can be due to the fact that E4T outputs often have very strong structural resemblance to the reference face image (very similar pose, size of face in the image, realistic face). It is known that DINO and CLIP-I metrics favor structure and semantics to original images respectively, instead of nuanced subject and style features. This is not necessarily something we want in the generation of a new stylized portrait of a face, with variation in pose, expression and style of a face being preferred in many applications.

\section*{Limitations}
Given the statistical nature of HyperNetwork prediction, some samples that are OOD for the HyperNetwork due to lighting, pose, or other reasons, can yield subotpimal results. Specifically, we identity three types of errors that can occur. There can be (1) a semantic directional error in the HyperNetwork's initial prediction which can yield erroneous semantic information of a subject (wrong eye color, wrong hair type, wrong gender, etc.) (2) incorrect subject detail capture during the fast finetuning phase, which yields samples that are close to the reference identity but not similar enough and (3) underfitting of both HyperNetwork and fast finetuning, which can yield low editability with respect to some styles.


\section*{Societal Impact}
\label{sec:societal}

This work aims to empower users with a tool for augmenting their creativity and ability to express themselves through creations in an intuitive manner. However, advanced methods for image generation can affect society in complex ways~\cite{saharia2022photorealistic}. Our proposed method inherits many possible concerns that affect this class of image generation, including altering sensitive personal characteristics such as skin color, age and gender, as well as reproducing unfair bias that can already be found in pre-trained model's training data. The underlying open source pre-trained model used in our work, Stable Diffusion, exhibits some of these concerns. All concerns related to our work have been present in the litany of recent personalization work, and the only augmented risk is that our method is more efficient and faster than previous work. In particular, we haven't found in our experiments any difference with respect to previous work on bias, or harmful content, and we have qualitatively found that our method works equally well across different ethnicities, ages, and other important personal characteristics. Nevertheless, future research in generative modeling and model personalization must continue investigating and revalidating these concerns.

\begin{table}[t]
\caption{\textbf{Comparisons with E4T.} This table presents a comparative analysis of our method against E4T for face identity preservation (Face Rec.), subject fidelity (DINO, CLIP-I), and prompt fidelity (CLIP-T). Our method demonstrates higher performance in face identity preservation and prompt fidelity, indicating a closer adherence to the original identity and text prompts.
\label{table:e4t_comparison}}
\centering
\resizebox{\columnwidth}{!}{
  \begin{tabular}{lccccc}
    \toprule
    Method & Face Rec. $\uparrow$ & DINO $\uparrow$ & CLIP-I $\uparrow$ & CLIP-T $\uparrow$\\
    \midrule
    Ours & \textbf{0.7076} & 0.528 & 0.628 & \textbf{0.284} \\
    E4T & 0.693 & \textbf{0.634} & \textbf{0.679} & 0.280 \\
    \bottomrule
  \end{tabular}
}
\end{table}

{
    \small
    \bibliographystyle{ieeenat_fullname}
    \bibliography{main}
}